\documentclass[conference]{IEEEtran}
\IEEEoverridecommandlockouts
\usepackage{cite}
\usepackage{amsmath,amssymb,amsfonts}
\usepackage{graphicx}
\usepackage{textcomp}
\usepackage{xcolor}
\usepackage{hhline} 

\usepackage{subfigure}
\usepackage{multirow}

\usepackage[ruled,linesnumbered]{algorithm2e}

\usepackage{lineno,hyperref}
\modulolinenumbers[5]

\def\BibTeX{{\rm B\kern-.05em{\sc i\kern-.025em b}\kern-.08em
    T\kern-.1667em\lower.7ex\hbox{E}\kern-.125emX}}
\begin{document}

\title{A Decision-Based Dynamic Ensemble Selection Method for Concept Drift\\
\thanks{FAPEAM - Funda\c{c}\~ao de Amparo \`a Pesquisa do Estado do Amazonas - Edital 009/2017 and Samsung Electronics of Amazonia Ltda, under the terms of Federal Law n$^o$ 8.387/1991, through agreement  n$^o$  003, signed with  ICOMP/UFAM.}
}

\author{\IEEEauthorblockN{R\'egis Ant\^onio Saraiva Albuquerque}
\IEEEauthorblockA{\textit{Federal University of Amazonas}\\
Manaus, Brazil \\
regis.albuquerque1@gmail.com}
\and
\IEEEauthorblockN{Albert Fran\c{c}a Josu\'a Costa}
\IEEEauthorblockA{\textit{Federal Institute of Amazonas} \\
Manaus, Brazil \\
albert.costa@ifam.edu.br}
\and
\IEEEauthorblockN{Eulanda Miranda dos Santos}
\IEEEauthorblockA{\textit{Federal University of Amazonas}\\
Manaus, Brazil \\
emsantos@icomp.ufam.edu.br}
\and
\IEEEauthorblockN{Robert Sabourin}
\IEEEauthorblockA{\textit{\'Ecole de Technologie Sup\'erieure} \\
Montreal, Canada \\
Robert.Sabourin@etsmtl.ca}
\and
\IEEEauthorblockN{Rafael Giusti}
\IEEEauthorblockA{\textit{Federal University of Amazonas}\\
Manaus, Brazil \\
rgiusti@icomp.ufam.edu.br}
}

\maketitle

\begin{abstract}
%Many machine learning practical applications are prone to concept drift, which may reduce the accuracy of classification systems significantly. This problem is especially important in big data applications, where data arrives in a stream, and patterns in the data are expected to evolve over time. In this context, methods using ensemble of classifiers are interesting due to the fact that these methods allow the design of strategies robust to cope with changes. 
We propose an online method for concept drift detection based on dynamic classifier ensemble selection. The proposed method generates a pool of ensembles by promoting diversity among classifier members and chooses expert ensembles according to global prequential accuracy values. Unlike current dynamic ensemble selection approaches that use only local knowledge to select the most competent ensemble for each instance, our method focuses on selection taking into account the decision space. Consequently, it is well adapted to the context of drift detection in data stream problems. The results of the experiments show that the proposed method attained the highest detection precision and the lowest number of false alarms, besides competitive classification accuracy rates, in artificial datasets representing different types of drifts. Moreover, it outperformed baselines in different real-problem datasets in terms of classification accuracy. %Additionally, it achieved lower detection delay and false detection rates, besides not presenting missing detection. %Additionally, we conducted experiments for visualizing diversity calculated on streams. These results indicated that diversity is affected by changes over time.% Therefore, this work shows that dynamic classifier ensemble selection performed on diverse populations of ensembles helps to improve the detection precision and the general accuracy of classification systems employed in problems with concept drift.
\end{abstract}


\section{Introduction}
Practical tasks, such as identification of customer preferences, Internet log analysis, among others, are examples of data stream problems. In this context, the so-called concept drift phenomenon may occur, since when data are continuously generated in streams, data and target concepts may change over time. Algorithms designed to deal with drift may be divided into two main groups: (1) online - when one instance is learned at a time upon arrival; and (2) block-based - when chunks of samples are presented from time to time \cite{Brzezinski2016}. Online methods are very useful in data stream environments, especially due to three main reasons:  samples arrive sequentially; data usually must be processed in high volumes at fast paces; and each data instance is read only once. 

Different categories of online methods are available in the literature. Drift detectors are common solutions. They focus on monitoring whether the class distribution is stable over time by observing the predictions of a classifier \cite{{BARROS2019213}}. Therefore, it is assumed that one instance is processed and its true label is available right after its classification. This scenario is described as an operational setting whose availability of the ground truth occurs in the next time step, as in food sales prediction \cite{Zliobaite2016}. 

Online learning ensembles have also been investigated as a strategy to tackle the drift problem. These methods update their knowledge base by adding, removing, or updating classifiers usually in a blind manner, i.e. they do not provide an explicit mechanism to
detect concept drift. However, there are also informed ensemble-based methods. In this case, the explicit drift detection mechanism mainly relies on employing an auxiliary drift detector \cite{BARROS2019213}, such as in Leveraging Bagging \cite{Bifet2010a}, which generates an ensemble of classifiers using a modified version of Online Bagging \cite{Oza2001} and employs ADWIN (Adaptive Window algorithm \cite{Bifet2007}) to detect drift. Moreover, some of these ensemble methods allow configurable drift detectors. In this case, most of the drift detectors available in the literature may be used internally \cite{BARROS2019213}. For instance, Boosting-like Online Learning Ensemble (BOLE) \cite{BARROS2019213}, which is a modified version of Online Boosting \cite{Oza2001}, uses a configurable drift detector, such as ADWIN or DDM (The Drift Detection Method) \cite{Gama2004}.

These previous works on informed online learning ensembles have two characteristics in common: 1) they encourage diversity only implicitly - by using algorithms such as the online versions of Bagging and Boosting; and 2) they generate only one ensemble of classifiers. However, by explicitly promoting diversity, one may better cope with drift and be more robust to false alarms, since it has been shown that different levels of ensemble diversity are required before and after a drift \cite{Minku2012} and that diversity changes over time \cite{Brzezinski2016}. Besides, it is possible to provide different levels of ensemble diversity over time by using a population of ensemble of classifiers instead of only one ensemble, which may also explicitly encourage diversity among the population members. These hypotheses are investigated in this paper.

We propose an ensemble-based method that allows configurable drift detectors. Here, however, a diverse population of ensembles is generated, whose members' diversity may range from low to high values, since the literature shows that diversity plays probably different roles depending on the drift type and frequency. Considering the fact that we propose an online method and that a class must be estimated for each unknown instance, we need to define a strategy to maximize the possibility of making the right decision when assigning a selected class. In our method, instead of pooling the decisions of every ensemble, we perform dynamic ensemble selection (DES) to elect a single ensemble that is probably the best qualified to predict a class to a given instance. The selection is called dynamic because an expert ensemble is defined for each unknown instance as it arrives on the fly. %In this paper, we intend to go one step further by generating a population of ensembles to maintain ensembles with different diversity levels and by performing dynamic ensemble selection (DES).

In the context of DES, most methods employed in concept drift problems are block-based \cite{ALMEIDA201867} and perform DES by defining local data regions (neighborhood) to estimate ensemble members' competence. However, the strategies used to generate pools of ensembles are mainly global data-oriented. The difference between local and global perspectives may prevent DES based on local competence to be highly effective \cite{CRUZ2018195}. Unlike these current works, the strategy employed in this paper for dynamic selection is decision-based, focused on selecting experts according to their accuracy calculated on the decision space. 

%The proposed method, called Dynamic Ensemble Selection for Drift Detection (DESDD), is divided into four phases: (1) diverse ensembles generation; (2) DES; (3) drift detection; and (4) drift reaction. First, a population of $c$ ensembles is created, where each ensemble is generated by an online algorithm and presents different diversity levels. Then, for each new instance to be classified, one ensemble is selected as an expert to assign a class to the new instance based on global prequential accuracy values. In the sequence, a configurable drift detector is used. If a change is observed, the system reacts to adapt to the new context and a new online process involving population generation and dynamic selection is started.  

Experiments were performed using artificial datasets, representing different types of drift, and real datasets. Two configurations of our method, called Dynamic Ensemble Selection for Drift Detection (DESDD), were investigated. Moreover, DESDD was compared to three baselines: DDM - which employs a single classifier; Leveraging Bagging - that works with one ensemble of classifiers; and DDD - which provides a population of ensembles, but performs static selection instead of DES. The results achieved on artificial datasets show that DESDD is more stable and more effective on detecting drifts than all the baselines, since the DESDD detector is triggered more often to indicate correct detections, attaining both lower false detection and miss detection rates as a consequence. The stability and effective capacity of drift detection provided by DESDD is also confirmed with the results obtained using real datasets, since DESDD outperformed all the baselines in terms of classification accuracy in these datasets. 

This paper is organized as follows. Related work is discussed in Section \ref{sec:related_work}. Thus, DESDD is detailed in Section \ref{sec:DESDD}. Experiments and results are summarized in Section \ref{sec:experiments}. Finally, conclusion and future work are discussed in Section \ref{sec:conclusion}.

\section{Related Work}
\label{sec:related_work}
Classifier ensembles are considered the most popular technique for handling concept drift both when employing change detection mechanisms and when working in a blind adaptation manner. Several ensemble solutions for online learning rely mainly on methods that implicitly encourage diversity during training, such as the modified versions of Online Bagging and Online Boosting \cite{Oza2001}. These two algorithms use a Poisson distribution to simulate their offline versions. For instance, Online Bagging updates ensemble components by presenting incrementally each instance to a component $k$ times, where $k$ is defined by the $Poisson(1)$ distribution, and assigns the final prediction by majority voting.

%In addition, 
%Even though DCS has been used in several ensemble methods for dealing with concept drift \cite{ALMEIDA201867}, very few authors have proposed methods using DES, and still fewer using diversity in the context of DES \cite{Gomes:2017}. Taking into account this scenario, works mentioned in this section are based on diversity or DES, for dealing with drifts in data stream problems by means of block or online learning.

% In \cite{Minku2010}, the authors created ensembles of Decision Trees using a version of Online Bagging they tailored to assign different values for the parameter $\lambda$ of $Poisson(\lambda)$ function. In this way, they were able to increase or reduce diversity by varying $\lambda$ values. Thus, they present a study to analyze whether or not diversity is an important classifier ensemble property in the context of online learning and drifts. It is important to note that Online Bagging is not composed of any specific approach to handle concept drift and does not perform DCS or DES. The authors pointed out the following conclusion: diversity plays an important role, which varies according to the type of concept drift and the time of change occurrences, i.e. high diversity is more important for severe drift, for other type of drift, its effect is very small.

One variant of Online Bagging is proposed in \cite{Bifet2010a}, called Leveraging Bagging. It adds more randomization to the original version by increasing resampling and using output detection codes. The first is conducted by including the parameter $\lambda$ to compute the $Poisson(\lambda)$ distribution. They set $\lambda=6$ to explicitly increase diversity. The second adjustment encourages diversity due to the reduction of experts' predictions correlation. Leveraging Bagging uses ADWIN as change detector, but it can be configured to use other methods \cite{BARROS2019213}. Summarizing, Leveraging Bagging generates a diverse ensemble of $n$ classifiers by applying the $Poisson(\lambda)$ function in a Test-Then-Train approach. In this scenario, the ensemble training is performed using data instances one at a time. When the true class of a test instance is discovered, this instance is subsequently considered as training sample. The hyperparameter $\lambda$ is fixed for all iterations and majority vote is used to predict classes of unknown examples. When ADWIN indicates a drift, the worst classifier member is removed and a new classifier is added to the ensemble. Since the whole ensemble of classifiers is used for making predictions, DES is not performed in Leveraging Bagging.

%ADWIN works using a window of recent observed items to monitor if there is a change in the average value inside the window. 
In order to analyze how diversity is affected by concept drift, as well as to study the possibility of using diversity as criterion for drift detection, Brzezinski and Stefanowski \cite{Brzezinski2016} adapted six diversity measures, widely used in static learning, to data streams requirements. They focused on showing visualizations of diversity on streams with various types of drifts and used diversity as criterion on the Page-Hinkley test for drift detection. They conducted experiments using Online Bagging to create several homogeneous ensembles and calculated diversity in three processing scenarios: blocks, incrementally and prequentially. Their results indicated that ensemble diversity visibly changes over time and some diversity measures, especially interrater agreement, are capable of detecting drifts as effectively as accuracy. Again, DES was not conducted. %In addition, diversity was not directly enforced when creating ensembles.%, since Online Bagging was not modified.

%More recently, Jackowski \cite{JACKOWSKI201823} advocated that the current diversity measures are not used to help dealing with drift because they cannot be applied when processing streaming data. According to this author, even the notion of diversity must be adapted to the context of streams of data, besides the proposal of new diversity measures especially designed for this kind of problem. Based on this premise, two new diversity measures and an error trend diversity driven ensemble generation method are proposed focusing on stream changes. Continuar.... 

A list and description of popular online ensemble methods for concept drift that work with configurable drift detectors is provided in \cite{BARROS2019213}. The methods discussed use online versions of either Bagging or Boosting and, except for the Dealing with Drifts (DDD) \cite{Minku2012}, they all generate only one ensemble of classifiers and do not perform any kind of selection. DDD also employs the variation of Online Bagging used in Leveraging Bagging \cite{Bifet2010a}, but it maintains four classifier ensembles with different levels of diversity, which is obtained with different values $\lambda$. It is important to mention that the values $\lambda$ used are fixed parameters. DDD works in two modes: prior to drift and after drift detection. In the first mode, two ensembles are created: one with low diversity and the other with high diversity, but only the low-diversity ensemble is used to assign labels to incoming instances. When the drift detector indicates a drift, DDD goes to the second mode. In this mode, two new ensembles are created, again with low and high diversity. Given that four ensembles are available, only the new high-diversity ensemble is not used to classify incoming instances. Weighted majority vote is employed as fusion function, where weights are defined according to ensembles prequential accuracy. Subsequently, DDD employs a heuristic to lead the system back to the first mode, where only two ensembles are dealt with and the new ensembles may replace the old ones. The authors indicate that DDD is proposed to better exploit the advantages of diversity, since it is able to maintain ensembles with different diversity levels. However, the static selection procedure used may be a limitation to achieve this objective.    

Differently from the works previously discussed, Almeida et al. \cite{ALMEIDA201867} propose a method (Dynamic Selection Based Drift Handler - Dynse) focused on DES designed to detect drift. Dynse works with a stream of batches and creates a classifier for each batch of data. Thus, when an unknown instance arrives, the set of $k$ nearest neighbors from a validation dataset surrounding the instance is defined and the most performing ensemble on this neighborhood is chosen to label it. In order to avoid the pool size to increase indefinitely, Dynse uses a pruning mechanism focused on diversity of region and concepts. Although the authors mention that the framework can be adapted to work as an online learning method, they did not perform the adaptation. Hence, this method may be assumed as a batch-based approach and, due to the neighborhood calculation, it may present the limitation related to local-driven dynamic selection \cite{CRUZ2018195}.

% \begin{figure*}%[!htp]
% \center
% \includegraphics[width=0.8\linewidth]{metodo_fases_en.png}
% \caption{The four phases of the proposed method: 1) diverse ensembles generation; 2) dynamic ensemble selection; 3) drift detection; and 4) drift reaction.} 
% \label{fig:metodo_proposto1}
% \end{figure*}

Our method processes incoming instances online as in \cite{Minku2012}, \cite{Bifet2010a}, and \cite{Brzezinski2016}. It employs the Online Bagging version used in \cite{Minku2012} and enforces more diversity by generating a population of diverse ensembles. Also, it performs DES based on prequential accuracy, i.e. it does not need to find the neighborhood of the unknown instances to estimate competence. Finally, DESDD also uses a configurable auxiliary drift detector. 

%DDM was proposed in \cite{Gama2004} and it defines two thresholds: warning and drift levels. The first level is reached if condition (1) is attained, while the drift level is achieved when condition (2) is satisfied. The values $p$ and $s$ represent the error rate of the learning algorithm and its standard deviation, respectively. Registers $p_{min}$ and $s_{min}$ are defined during the training phase, and are updated if after each incoming example $x_t$, the current register $p_t + s_t$ is lower than $p_{min} + s_{min}$.

%\begin{equation}
 % p_t + s_t >= p_{min} + 2*s_{min} 
 % \label{eq:warning}
%\end{equation}

%\begin{equation}
 % p_t + s_t >= p_{min} + 3*s_{min} 
 % \label{eq:drift}
%\end{equation}

%Our proposed method is described in the next section.

\section{Proposed Method}
\label{sec:DESDD}
%In this section, we describe our method DESDD. It is an online approach employed in the context of the Test-Then-Train scenario. Precisely, the classifier ensembles’ training is performed using data instances sequentially, i.e. one at a time. In addition, when a test instance true class identity is known, this instance is subsequently considered as training sample. As a consequence, the classifier ensembles’ training must be updated throughout time. 
In order to simplify the description of the proposed method, we use the following notation. Let $x_t \in \mathbf{R}^{d}$ be a $d$-dimensional labeled instance observed in time step $t$ whose real class is $y_t$, from a classification problem with a finite set of class labels. % $\Upsilon=\{y_1, y_2,\dots,y_l\}$
Instances are observed one at a time, and the time intervals are not necessarily equally spaced. Given that an unknown instance $x_{t+1} \in \mathbf{R}^{d}$ arrives at time step $t+1$, the objective of DESDD is to assign a label $y^{*}$ to $x_{t+1}$. After this decision, the true label $y_{t+1}$  becomes available. Then, the pair $(x_{t+1},y_{t+1})$ is used to update the population of ensembles.

DESDD is divided into four phases: (1) diverse ensembles generation; (2) DES; (3) drift detection; and (4) drift reaction. First, by employing a configurable online ensemble creation algorithm, a population $P=\lbrace C_1, C_2, ..., C_c \rbrace$ composed of $c$ ensembles, each with $n$ components and different levels of diversity, is generated. Then, for each new instance $x_{t+1}$ to be predicted, the ensemble candidate ($C_j^*$) with the highest current prequential accuracy is selected as an expert to assign a label $y^*$ to $x_{t+1}$. As soon as the true class $y_{t+1}$ becomes available, DESDD goes into the drift detection phase, where a configurable detector is used. If no drift is detected, the ensembles in $P$ are updated by training with $(x_{t+1},y_{t+1})$. If a drift is detected, the drift reaction phase is reached. In this case, the entire population is replaced and a new process of diverse ensemble population generation and dynamic ensemble selection begins. We describe the details of DESDD phases in the next subsections. %Figure ~\ref{fig:metodo_proposto1} illustrates our method.

\subsection{Diverse Ensembles Population Generation}

The first phase involves the online generation of a population of diverse ensembles to allow decision-based dynamic selection. The strategy used to explicitly enforce diversity when creating ensembles depends on the online ensemble creation algorithm employed. It is possible to observe that this phase of DESDD may be undertaken using any online ensemble generation method that explicitly enforces diversity, such as the modified versions of Online Bagging used in \cite{Minku2012} and Online Boosting used in \cite{BARROS2019213}. As previously mentioned, both algorithms have the same parameter $\lambda$ that determines the level of explicit diversity inserted when creating an ensemble. In this work, we employ Online Bagging. Taking into account that a diverse classifier ensemble is created by varying the values $\lambda$, we define the range for the values $\lambda$  (minimum and maximum values $\lambda$) as a parameter. Thus, each DESDD iteration involves assigning a numeric value in the range $[\lambda_{min}, \lambda_{max}]$ for each ensemble member of $P$ randomly. 

Besides the online learning algorithm and the minimum and maximum values of $\lambda$, the following parameters must be set at the first phase of DESDD: 1) size of the population $P$; 2) number $n$ of components of each ensemble; and 3) types of components -- the ensembles may be either homogeneous or heterogeneous. After creating the population $P$, when $x_{t+1}$ arrives, DESDD reaches the dynamic selection phase. It is important to point out that this process follows the Test-Then-Train technique. However, without loss of generality, it is also possible to perform initial training with a batch of labeled instances and then follow the Test-Then-Train learning process.

%Thus, after creating the ensemble population $P$, when $x_{t+1}$ arrives, DESDD attains the dynamic selection phase.

\subsection{Dynamic Ensemble Selection}

The goal of the selection phase of DESDD is choosing an expert ensemble $C_j^*$ to assign a label for each unknown instance $x_{t+1}$. The expert is chosen as the ensemble with highest prequential accuracy up to instance $x_t$, that is, its prequential accuracy is not a local measure. Such ensemble is expected to be the most likely to assign a correct label to $x_{t+1}$. The prequencial accuracy (fusion of \textbf{pre}dictive and se\textbf{quential}), also called Interleaved Test-Then-Train, is the evaluation metric more often used in data stream mining. It is the average accuracy of predicted examples prior to the example being learned, and it is computed at time step $t$ by:

\begin{equation}
  Acc = 
	\begin{cases}  
  	  acc_{ex}(t), & \text{if}\ t=f \\
      acc(t-1) + \frac{acc_{ex}(t) - acc(t-1)}{t-f+1}, & \text{otherwise}
  	\end{cases}
  \label{eq:PA}
\end{equation}

In this equation, $acc_{ex}$ is 0 when the prediction of the current training example  is incorrect and 1 otherwise. The term $f$ indicates the first time step.% Once the true class of $x_{t+1}$ has been defined, DESDD follows to the drift detection phase.

%Algorithm ~\ref{algoritmoselecao} summarizes the dynamic selection phase of DESDD. 
%The input parameters for this algorithm are as follows: the population of classifier ensembles $P$ created in the previous phase; the amount of ensembles $c$; and the instance to be classified $x_{t+1}$. First, the prequential accuracies of classifier ensembles $ACC$ are calculated up to instance $x_{t}$. In the sequence, a ranking function is employed (line 4) to put classifier ensembles in a descending order of accuracy. Then, the ensemble with the highest accuracy value is chosen as the expert solution $C_j^*$ (line 5). Finally, the $x_{t+1}$’s label is defined by majority voting among classifier members of $C_j^*$. Once the true class of $x_{t+1}$ has been defined, DESDD follows to the drift detection phase.

%\begin{algorithm}
 %   \caption{Dynamic selection phase based on prequential accuracy}
 %   \label{algoritmoselecao}
 %   \begin{algorithmic}[1] % The number tells where the line numbering should start
  %  	\INPUT $P$; $P$ size $c$; unknown instance $x_{t+1}$; and sliding window $W$.
   %     \FOR{ ($j \gets 0$; j < c; j++ ) } 
    %        \STATE $ACC(C_j) \gets ACC.calculate(C_j)$
     %       \ENDFOR
     %   \STATE $Ranking \gets %FindACCRanking(P, ACC, c) $  
     %   \STATE $C_j^* \gets ACC.select(Ranking) $
      %  \STATE Return $Majorityvoting(C_j^*, x_{t+1})$
   % \end{algorithmic}
%\end{algorithm}

\subsection{Drift Detection}

In this third phase, a drift detector based on error monitoring is employed. Here, the prediction $y^{*}$, provided by the $C_j^*$ chosen in the previous phase, is compared to the true label $y_{t+1}$ of $x_{t+1}$ in order to indicate whether or not the prediction was correct. Then, this information is used in an error monitoring process to indicate the occurrence of a drift. Since our method is configurable with a drift detector, several strategies, ranging from traditional to newer detectors can used. In \cite{BARROS2019213}, a large-scale comparison of ten drift detectors is performed in the context of ensemble algorithms configurable with detectors. The authors indicate that no single drift detector may be deemed to be the best strategy in all situations. Based on this observation, although DESDD can be configured to use any detector investigated in \cite{BARROS2019213}, in this work we employ the following two drift detectors: DDM and ADWIN. The former, is a traditional drift detector based on error monitoring. The latter, is employed in conjunction with Leveraging Bagging, which is a baseline used in our experiments. 

%used in the first phase of DESDD. However, other methods such as EDDM and AUE2 may also be used. % interesting options. %In addition, since DESDD was designed using the MOA framework, any supervised drift detector available in this public framework may be used in this phase of DESDD, even new proposed detectors. 

%The way each drift detector works is described in section \ref{sec:related_work}. 
%Especially noteworthy is the fact that, considering the different operation mode of each drift detector, the third phase of DESDD may be tailored to work according to these detectors. DDM defines two thresholds: warning level and drift level. Then, all samples dealt with between warning and drift levels are stored to be used to update classifiers after a drift detection. In terms of ADWIN, the process is the traditional Test-Then-Train scenario. These two scenarios are tested in this work. 

\subsection{Drift Reaction}
\label{subsec:drift_reaction}
For the last phase of DESDD, there are also different possible scenarios. In this work, the drift detector monitors each ensemble in $P$. When a drift is indicated, a new population of ensembles is created, i.e. the whole current population is discarded. The strategy used to generate the new population of ensembles after a drift detection depends on the drift detector employed. Considering the two detectors investigated in this work, two different training processes are conducted. When DDM is the drift detector, ensembles training is performed using instances stored by DDM during its warning level and drift level interval. In terms of ADWIN, training follows the traditional Test-Then-Train strategy.

\section{Experiments and Results}
\label{sec:experiments}
Our experiments were broken down into three main series. In the first, we investigate the behavior of the decision-based dynamic selection strategy employed by DESDD related to diversity variation in the presence of different types of drift. After that, the two versions of DESDD obtained by using DDM and ADWIN as drift detectors are compared. Then, in the third series, DESDD is compared to three baselines: (1) DDD; (2) DDM; and (3) Leveraging Bagging. These baselines were chosen mainly because DDD is a method that works with a population of ensembles (4 ensembles) and a configurable drift detector; DDM is a component of DESDD; and Leveraging Bagging was shown in \cite{BARROS2019213} to be competitive with the best ensemble algorithms for drift detection, especially when using Hoeffding Tree as classifier component, which is the base classifier used in our experiments. First, however, the datasets investigated are detailed.

\subsection{Datasets}
The experiments are conducted using ten artificial datasets (representing gradual and abrupt drifts with and without noise) and four real datasets. All artificial datasets are balanced, publicly available in the MOA framework, contain 10,000 instances, and were generated to present a drift at each block of 1,000 instances, i.e. they present nine concept drifts. For datasets with noise, they contain 10\% examples with class noise. The artificial datasets allow us to evaluate the methods in terms of detection rate, detection delay and miss detection, besides prequential accuracy. The datasets are: 

%datasets divided into two classes, their instances are represented by only two features and each 1000 instances represent a concept. 
\begin{itemize}
\item \textbf{Agrawal}: Four datasets were created with Agrawal generator based on all 10 functions available, two presenting gradual drift and two presenting abrupt drift - Agrawal Gradual; Agrawal Gradual with Noise; Agrawal Abrupt; and Agrawal Abrupt with Noise. The datasets are represented by nine features and divided into two classes. 
\item \textbf{Gauss}: Its instances are represented by two features and divided according to two Gaussian distributions, i.e. two classes. This dataset presents abrupt concept drifts, noise, and reversal of the classes after every drift. 
\item \textbf{SEA}: SEA generator based on all functions available was used
to create four datasets - SEA Gradual; SEA Gradual with Noise; SEA Abrupt; and SEA Abrupt with Noise. They contain 3 attributes and 3 classes.
\item \textbf{Sine1}: It is a dataset composed of instances generated according to a sine equation. This dataset presents gradual concept drifts, is noise-free and, after every drift, classes are reversed.
\end{itemize}

The real datasets only allow the analysis of the investigated methods regarding accuracy, since it is not possible to know when a drift starts to occur, which type of drift is present or even if there is drift in the data. This is a common characteristic in real applications. The following real datasets are investigated. All of them have been previously used in works dealing with concept drift detection.
\begin{itemize}
\item \textbf{Forest Covertype}: This dataset contains 581,012 instances divided into 7 classes representing forest cover types. A set of 54 features describes each instance.
\item \textbf{KDD Cup 1999}: It is composed of 489,844 instances divided into two classes, which represent several network intrusion problems simulated in a military network. Each instance is described by 41 features.
\item \textbf{Poker-Hand}: It contains 829,201 instances, 10 classes and its feature vector is composed of 11 attributes.
\item \textbf{SPAM}: It contains 9,324 instances divided into spam and ham classes and is represented by 499 binary features. 
\end{itemize}

\subsection{Base Classifiers}
Ensembles may be composed of identical or nonidentical base classifiers. Since the three baselines investigated in this work use Hoeffding Tree as base learning algorithm, in this work we generate homogeneous classifier ensembles composed of Hoeffding Tree, which is an online learning algorithm for large volumes of data generated in stream.
%were used in our initial experiments. Homogeneous ensembles were composed of Hoeffding Tree classifiers, which is an online learning algorithm for large volumes of data generated in stream proposed in \cite{Domingos2000}. In addition, most baselines employ Hoeffding Tree as base classifiers as well. For the heterogeneous ensembles, 
%These ensembles are composed of online versions of the following learning algorithms: Hoeffding Tree, kNN, Random Hoeffding Tree, Perceptron and NaiveBayes. This choice is focused on generating heterogeneous ensembles with base classifiers whose learning process employs different approaches.  

We have conducted experiments for evaluating the impact of the number of ensemble members of the population by varying the amount of members. Our results indicated a population size of 11 ensembles as ideal size. In addition, we have compared different ensemble sizes, between 10 and 25 components. The results showed no significant differences in terms of prequential accuracy by varying ensembles sizes. Thus,  the number of components of each ensemble was defined as 10, i.e. the smallest tested size. 

Besides population and ensembles sizes, the range for the values $\lambda$ used to generate the population of ensembles was also defined. The values used were: $\lambda_{min}=0.001$ and $\lambda_{max}=100$. This range has been expanded from the study presented in \cite{Minku2012}. The same range was used for all datasets. It is important to mention that, in all series of experiments, simple majority voting was employed as fusion function. 
\subsection{Analyzing Dynamic Selection vs. Diversity Variation}
\label{sec:analysis}
This section presents an analysis of the dynamic selection process conducted in the second phase of DESDD. We aim at answering the following question: how would the selection process behave considering a diverse population of ensembles in the presence of different types of drift? In order to understand such a behavior, this analysis is conducted using the population of ensembles that is generated at the first phase of DESDD. However, rather than randomly assign a numeric value in the range $[\lambda_{min}, \lambda_{max}]$ to each ensemble member of the population $P$, we have fixed the values $\lambda$ for the population of 11 ensembles to better monitor each one. Therefore, the same values were employed at each iteration: 0.001, 0.005, 0.01, 0.05, 0.1, 0.5, 1, 5, 10, 50, 100. In this first series of experiments, DDM is used as configurable drift detector. Moreover, only artificial datasets are investigated.

\begin{figure}
  \centering
  \subfigure[SEA Gradual]{\includegraphics[width=0.24\textwidth]{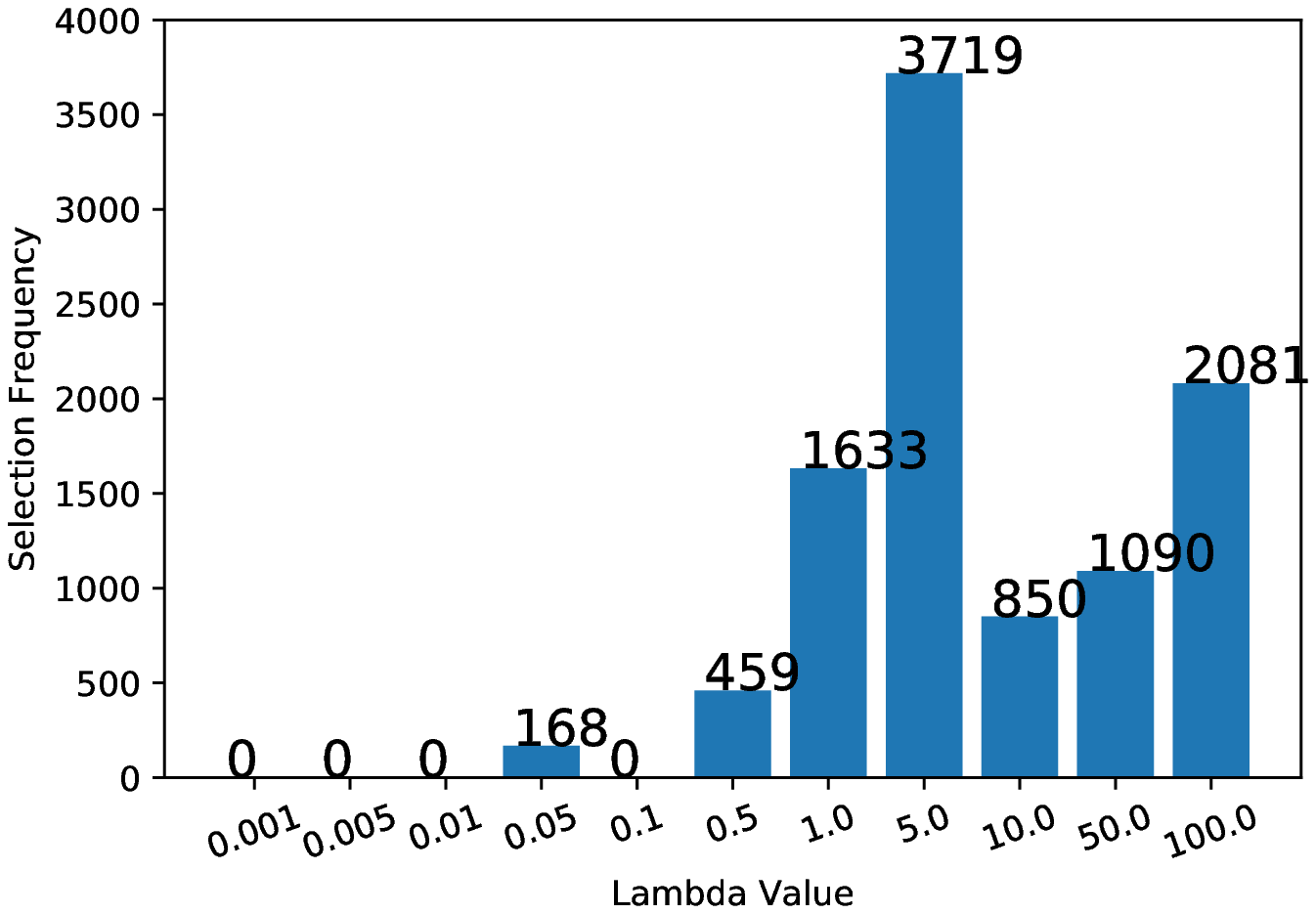}\label{SEA_Histogram}}
  \subfigure[SEA Gradual]{\includegraphics[width=0.24\textwidth]{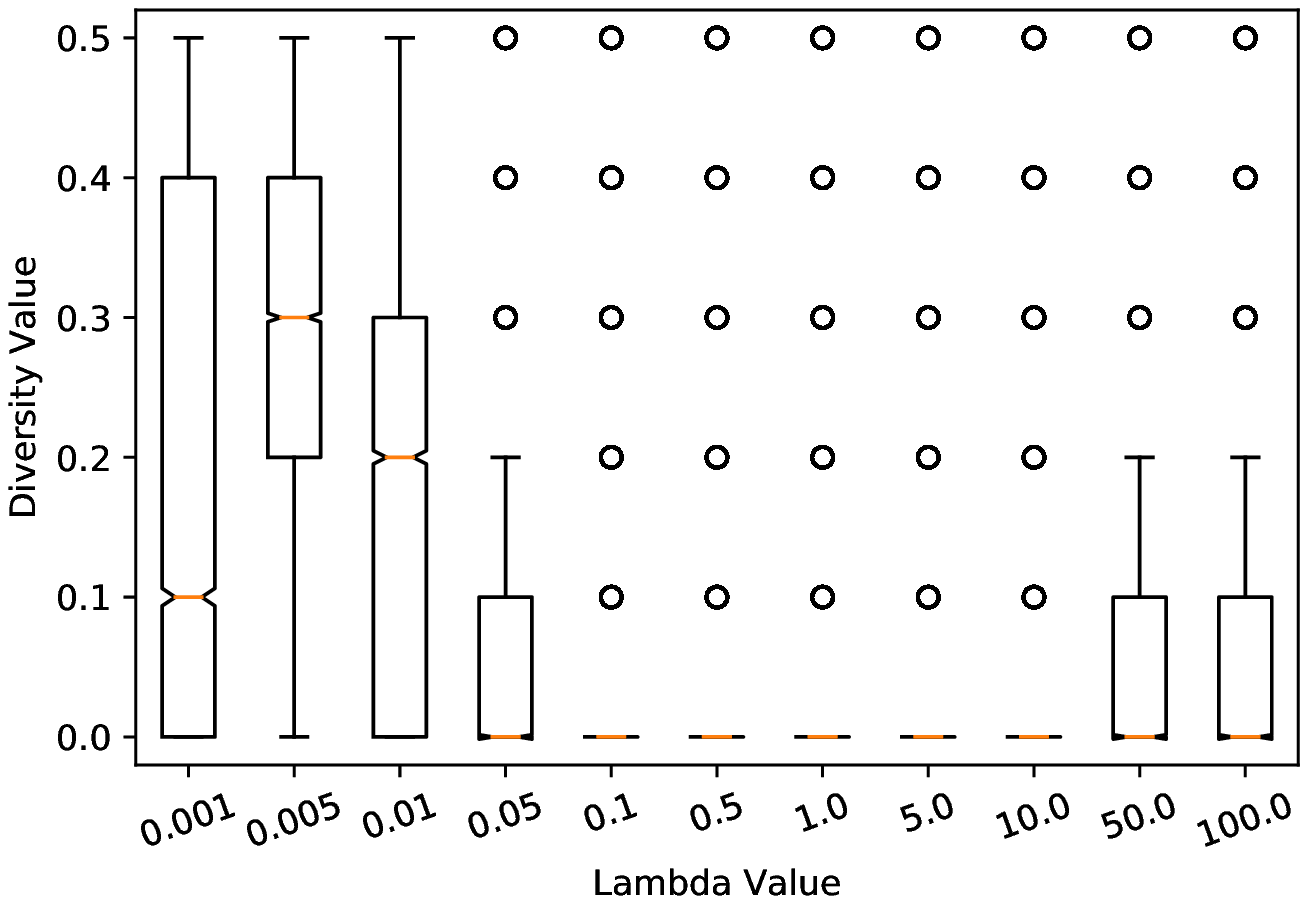}\label{SEA_Dispersion}}
  \subfigure[Agrawal Abrupt]{\includegraphics[width=0.24\textwidth]{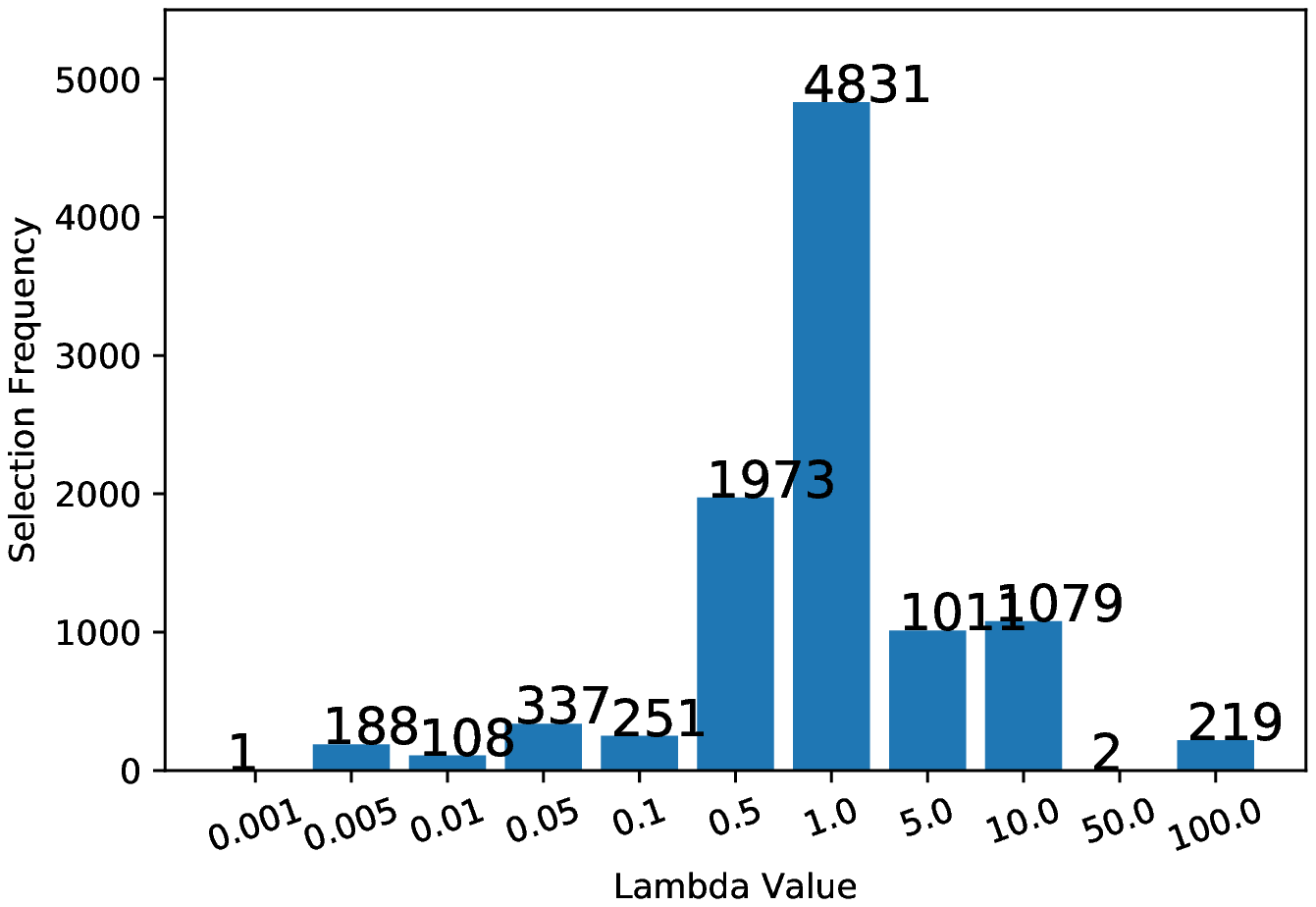}\label{Agrawal_Abrupt_Histogram}}
  \subfigure[Agrawal Abrupt]{\includegraphics[width=0.24\textwidth]{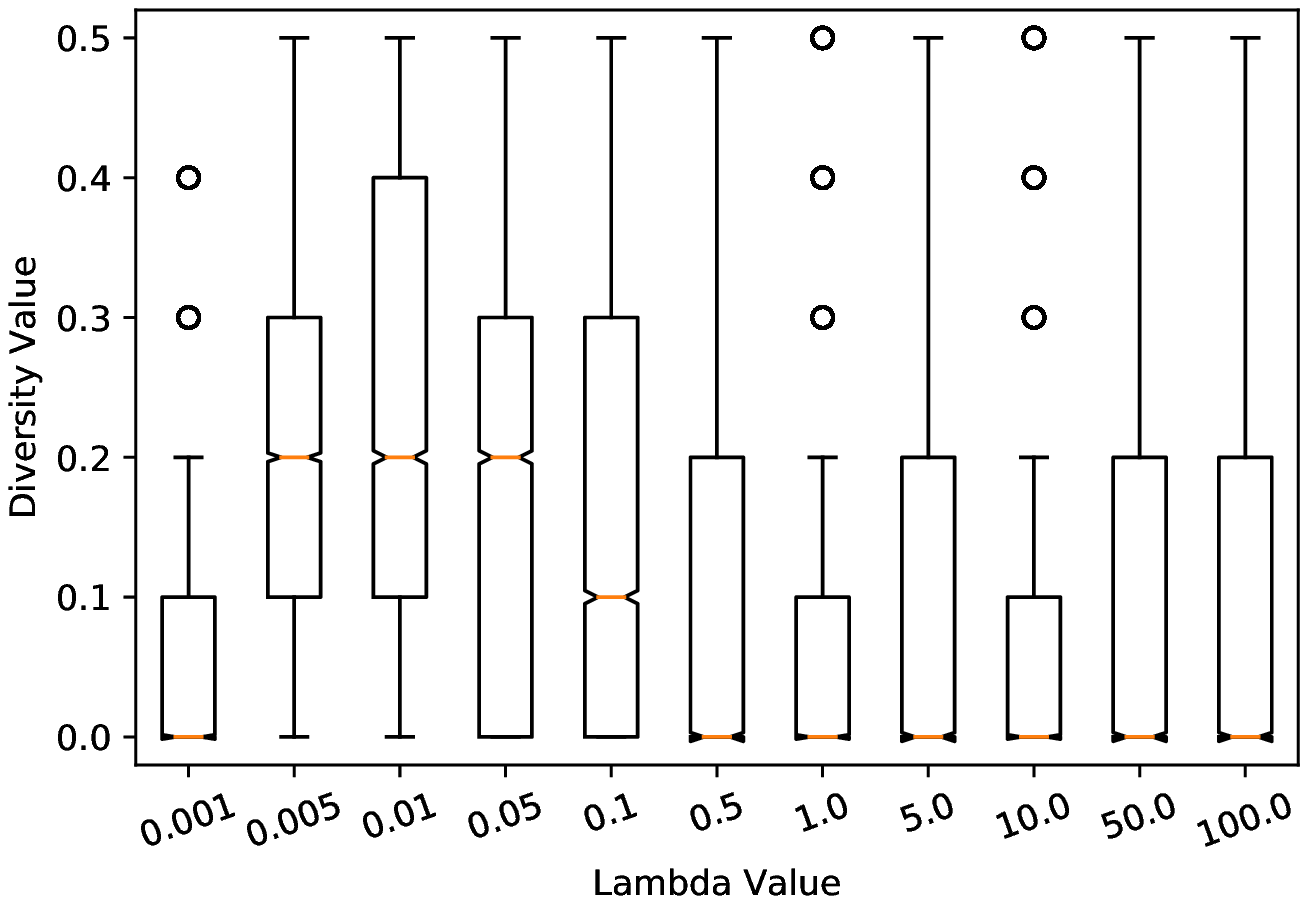}\label{Agrawal_Abrupt_Dispersion}}
   \subfigure[Gauss]{\includegraphics[width=0.24\textwidth]{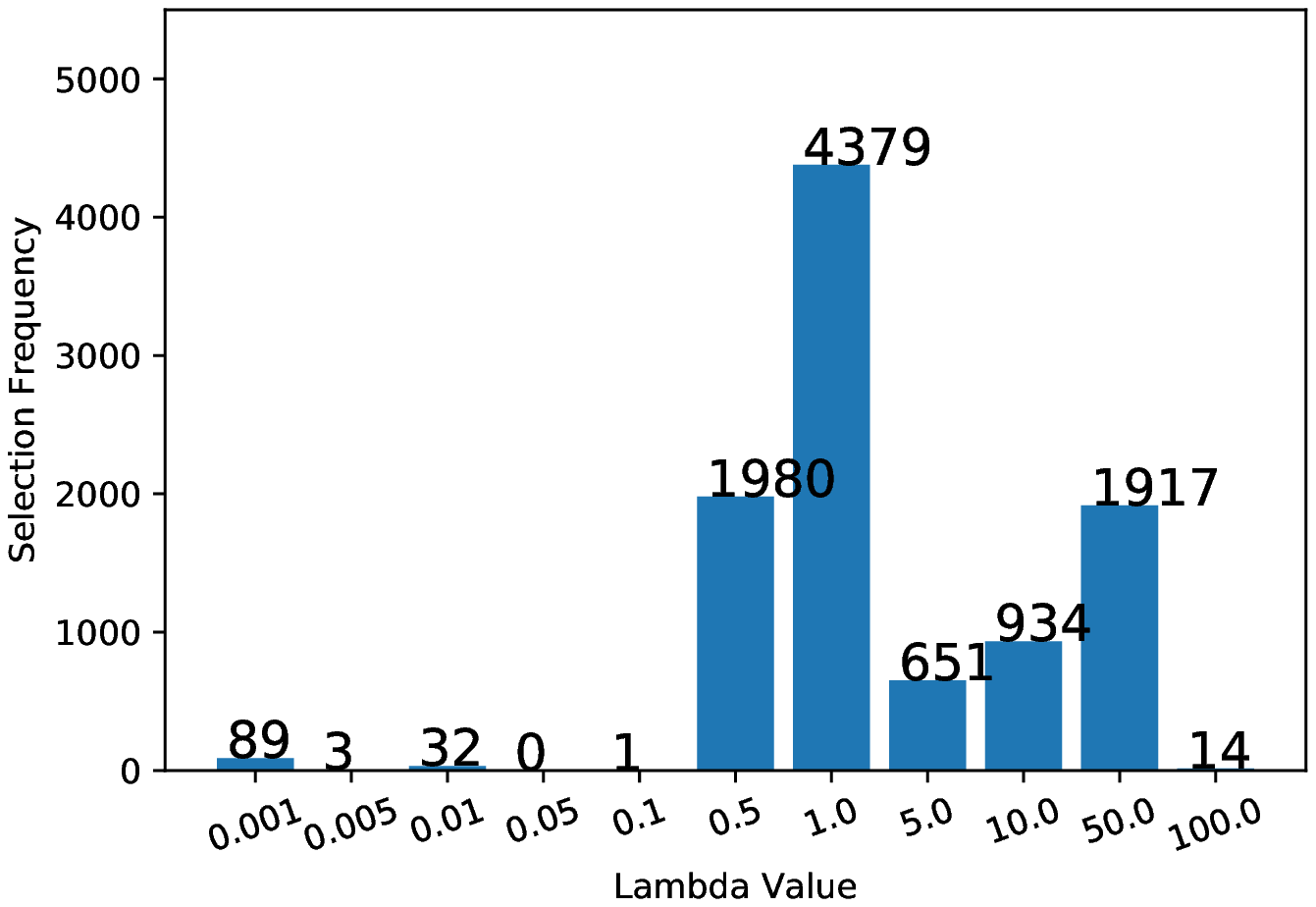}\label{Gauss_Histogram}}
   \subfigure[Gauss]{\includegraphics[width=0.24\textwidth]{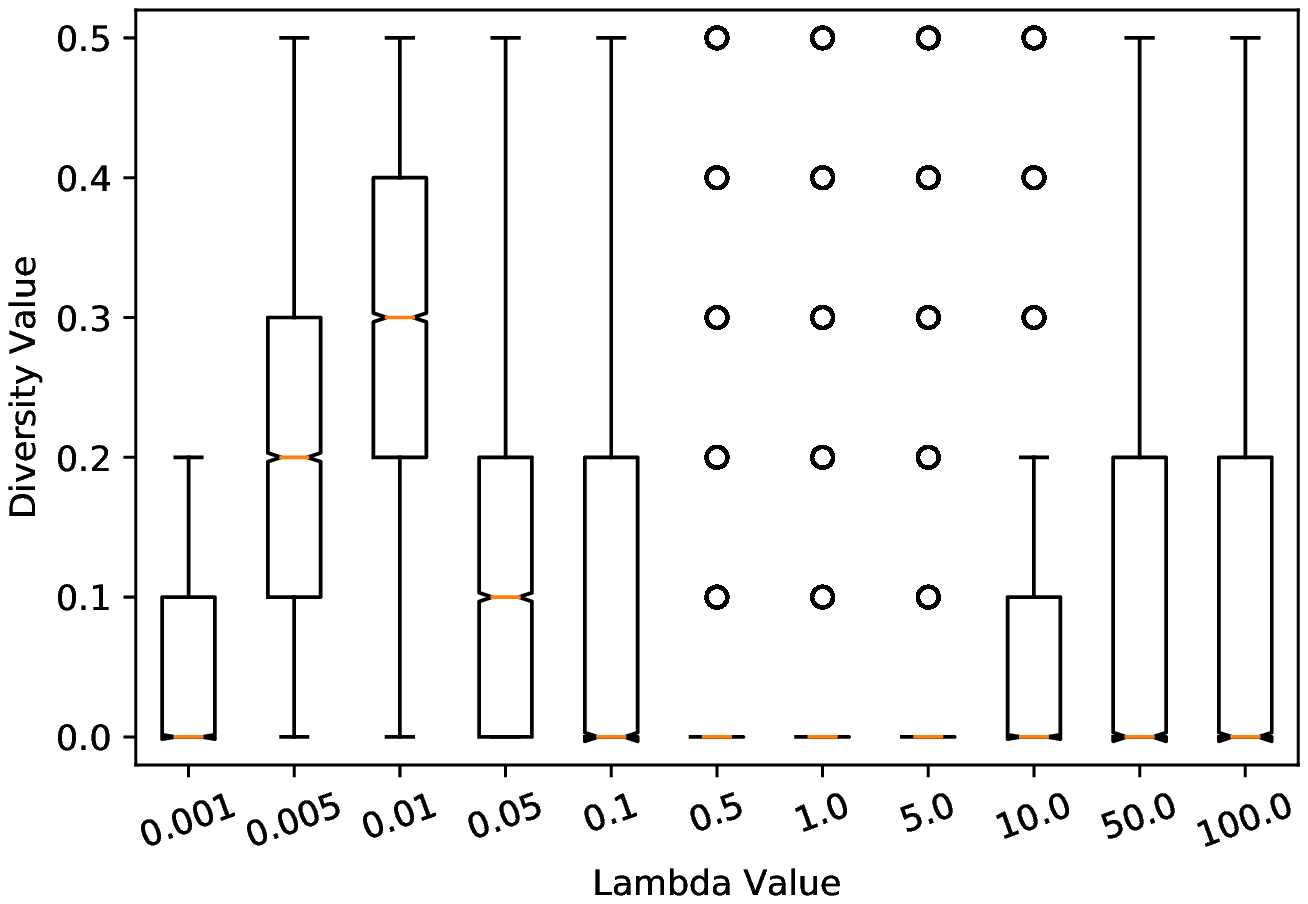}\label{Gauss_Dispersion}}
  \caption{Histograms of the frequency of selection and the distribution of diversity for each candidate ensemble on SEA Gradual (\ref{SEA_Histogram}, \ref{SEA_Dispersion}), Agrawal Abrupt (\ref{Agrawal_Abrupt_Histogram},\ref{Agrawal_Abrupt_Dispersion}) and Gauss (\ref{Gauss_Histogram}, \ref{Gauss_Dispersion}) datasets. DESDD is performed with fixed values $\lambda$ on these examples.}
  \label{fig:histogram_gradual}
\end{figure}

Figure \ref{fig:histogram_gradual} shows two types of information related to each candidate ensemble: 1) histograms of the frequency of selection and 2) the distribution of diversity. Here, diversity is measured as ambiguity. The classification ambiguity is an unsupervised dissimilarity measure defined for each candidate ensemble $C_j$ in terms of its component $c_i$ as:

\begin{equation}
\label{eq:ambiguidade}
  a_i(x_{t+1}) = 
	\begin{cases}  
  	  0, & \text{if}\ y_i=y_j \\
      1, & \text{otherwise}
  	\end{cases}
\end{equation}

where $a_i$ is the ambiguity, $y_i$ is the class assigned by component $c_i$ to instance $x_{t+1}$, and $y_j$ is $C_j$'s output. The final ambiguity for ensemble $C_j$ is:

\begin{equation}
  div = \frac{ 1 }{n} \sum\limits_{i \in C_j} a_i(x_{t+1})
\end{equation}

Due to lack of space, it is not possible to present histograms and distribution of diversity for all artificial datasets investigated in this paper. Hence, we show in Figure \ref{fig:histogram_gradual} three datasets only: SEA Gradual, Agrawal Abrupt and Gauss (abrupt drift with noise). Even so, some observations can be made considering all datasets investigated. First, although it can be clearly observed in our experiments that the ensemble with $\lambda=1.0$ is most often selected as expert (8 datasets out of a total of 10 datasets), the selection process presents higher variation for abrupt drift problems (figures \ref{Agrawal_Abrupt_Histogram} and \ref{Gauss_Histogram}) than for gradual drift problems (\ref{SEA_Histogram}). For gradual drift problems, the selection is more concentrated on ensembles whose values $\lambda$ are higher than $0.1$. Although in abrupt drift datasets the selection frequency is also higher for $\lambda > 0.1$,  almost all candidate ensembles are selected through the iterations. We may, therefore, indicate that a larger range for values $\lambda$ is more useful in problems with abrupt changes. It is important to observe that DES assures a high variance of expert selection. Second, noise does not play a key role on changing this behavior (Figure \ref{Gauss_Histogram}, which shows the results in a dataset with abrupt drifts and noise). Finally, the relation between $\lambda$ and diversity cannot be clearly observed. However, diversity is higher when $\lambda < 0.1$. In general, the highest diversity rate was attained by the ensemble with $\lambda=0.01$ (7 datasets out of the 10 datasets). The diversity behavior appears to be problem-dependent.

\subsection{Comparison of Different DESDD Versions}

In this second series of experiments, two DESDD versions are investigated by varying the drift detector: DDM and ADWIN. Again, only the artificial datasets were used in these experiments. It is important to mention that DESDD is employed as described in Section \ref{sec:DESDD}, i.e. each iteration involves randomly assigning a numeric value in the range $[\lambda_{min}, \lambda_{max}]$ to each ensemble member of the population $P$. Mean and standard deviation values of 30 replications obtained by each version of DESDD are shown in Table \ref{tab_metodos_reacao_reais} for all 10 datasets. The best result for each dataset is in bold. %This table also shows the average execution time of each version measured in seconds. 

% MÉTODO e MEDIDAS
\begin{table}[htbp]
\caption{Mean and standard deviation values of accuracy obtained on 30 replications comparing two versions of DESDD.} %Values in bold indicate the highest accuracy rate.} %, as well as execution time, measured in seconds, 

\begin{center}
%\begin{small}
\begin{tabular}{|l|c|c|} 
	\hline
	\textbf{Dataset} & {\textbf{DESDD-ADWIN}} & {\textbf{DESDD-DDM}}  \\  \hline
   % \cline{2-5}
	Agrawal Abrupt   & 76.51 (0.48) & \textbf{81.00} (1.45) \\ \hline
	Agrawal Abrupt Noise  & 73.34 (1.16) & \textbf{81.07} (1.17) \\ \hline
	Agrawal Gradual  & 73.15 (0.50) & \textbf{77.56} (1.37) \\ \hline
	Agrawal Gradual Noise & 70.97 (0.91) & \textbf{76.80} (1.71) \\ \hline
	Gauss   & 72.71 (0.85) & \textbf{86.04} (0.44) \\ \hline
	SEA Abrupt & 83.02 (2.23) & \textbf{93.24} (0.24) \\ \hline
	SEA Abrupt Noise & 68.55 (1.00) & \textbf{83.19} (0.43) \\ \hline
	SEA Gradual  & 82.23 (2.17) & \textbf{92.17} (0.29) \\ \hline
	SEA Gradual Noise & 68.17 (1.09) & \textbf{83.37} (0.59) \\ \hline
	Sine1 & 76.97 (1.15) & \textbf{93.04} (0.23) \\ \hline
\end{tabular}
%\end{small}
\label{tab_metodos_reacao_reais}
\end{center} 
\end{table}

% MÉTODO e MEDIDAS - BASES SINTÉTICAS
\begin{table*}[htbp]
\caption{Accuracy and execution time attained. Values in bold indicate the highest accuracy, while (*) indicates that a method is significantly better than all the others.}
\begin{center}
%\begin{small}
\begin{tabular}{|l|l|c|l|c|l|c|l|c|} 
	\hline
	 & \multicolumn{2}{c|}{\textbf{DESDD}} & \multicolumn{2}{c|}{\textbf{DDD}} & \multicolumn{2}{c|}{\textbf{DDM}} & \multicolumn{2}{c|}{\textbf{LB}}  \\  
    \cline{2-9}

	%\hhline{|~|-|-|-|-|-|-|-|}\noalign{\vspace{0pt}} 

	\multirow{-2}{*}{\textbf{Dataset}} & \textbf{Acc} & \textbf{Time} & \textbf{Acc} & \textbf{Time} & \textbf{Acc} & \textbf{Time}  & \textbf{Acc} & \textbf{Time} \\ \hline
	Agrawal Abrupt  & 83.41 ( 4.54 ) &  23.13 & 81.94 ( 4.24 ) &  2.8 & \textbf{84.13}* ( 5.2 ) &  0.09 & 81.22 ( 5.06 ) &  1.87 \\ \hline 
     Agrawal Abrupt Noise  & 79.22 ( 5.04 ) &  20.3 & \textbf{80.01} ( 4.86 ) &  2.18 & 79.96 ( 4.36 ) &  0.1 & 77.92 ( 4.63 ) &  1.54 \\ \hline 
     Agrawal Gradual  & 79.69 ( 4.16 ) &  24.8 & \textbf{79.76} ( 3.97 ) &  1.98 & 79.13 ( 5.03 ) &  0.08 & 78.14 ( 4.82 ) &  1.74 \\ \hline 
     Agrawal Gradual Noise  & 76.85 ( 4.48 ) &  22.09 & 78.37 ( 4.86 ) &  2.11 & \textbf{78.61} ( 4.35 ) &  0.07 & 73.92 ( 4.2 ) &  1.77 \\ \hline 
     Gauss  & 86.55 ( 1.13 ) &  5.27 & \textbf{86.63} ( 1.09 ) &  1.03 & 80.77 ( 3.25 ) &  0.03 & 76.55 ( 4.64 ) &  0.43 \\ \hline 
     SEA Abrupt  & \textbf{93.77}* ( 0.97 ) &  4.99 & 92.94 ( 1.03 ) &  0.9 & 92.92 ( 0.86 ) &  0.04 & 93.13 ( 0.61 ) &  0.32 \\ \hline 
     SEA Abrupt Noise  & 83.87 ( 0.65 ) &  7.88 & 81.87 ( 0.6 ) &  1.67 & \textbf{84.07} ( 0.83 ) &  0.04 & 83.46 ( 0.52 ) &  0.28 \\ \hline 
     SEA Gradual  & 92.46 ( 0.67 ) &  5.19 & 92.35 ( 0.88 ) &  0.64 & 92.07 ( 0.63 ) &  0.04 & \textbf{92.65} ( 0.49 ) &  0.28 \\ \hline 
     SEA Gradual Noise  & 83.17 ( 0.49 ) &  7.85 & 82.37 ( 0.79 ) &  1.12 & \textbf{83.8}* ( 0.79 ) &  0.04 & 83 ( 0.42 ) &  0.28 \\ \hline 
     Sine1  & \textbf{93.37}* ( 0.44 ) &  3.91 & 86.79 ( 0.73 ) &  1.87 & 89.31 ( 0.68 ) &  0.04 & 85.06 ( 2.49 ) &  0.56 \\ \hline
    Forest Covertype  & \textbf{93.28}* ( 2.49 ) &  1820.59 & 84.77 ( 1.7 ) &  555.93 & 87.45 ( 2.37 ) &  23.04 & 91.17 ( 2.72 ) &  146.34 \\ \hline 
    KDDCup  & \textbf{99.97} ( 0.01 ) &  496.94 & 99.96 ( 0.03 ) &  194.66 & 99.96 ( 0.01 ) &  7.9 & 99.97 ( 0.01 ) &  54.08 \\ \hline 
    Poker-Hand  & \textbf{91.06}* ( 0.85 ) &  1040.77 & 78.06 ( 0.69 ) &  254.38 & 73.68 ( 1.52 ) &  5.73 & 86.1 ( 0.77 ) &  49.82 \\ \hline 
    Spam  & \textbf{96.26}* ( 0.89 ) &  221.67 & 90.54 ( 2.51 ) &  68.65 & 90.89 ( 2.1 ) &  2.82 & 94.31 ( 1.96 ) &  20.26 \\ \hline
\end{tabular}
%\end{small}
\label{tab_metodos_comp_sin}
\end{center} 
\end{table*}

The results shown in Table \ref{tab_metodos_reacao_reais} allow us to observe that the two versions of DESDD attained very different accuracy rates. DESDD-DDM was significantly better than DESDD-ADWIN. It is important to note in this table that DESDD operating with ADWIN is more affected by noise than its version with DDM. For instance, the reduction of DESDD-ADWIN accuracy in the SEA Abrupt dataset was $17\%$ when noise was added - if we compare accuracy rates achieved in SEA Abrupt and SEA Abrupt with Noise datasets, while this difference is only $10\%$ for DESDD-DDM. This behavior is observed in all datasets with noise. Moreover, since the whole population of ensembles is replaced after a drift detection, DDM may present faster recovery from drifts due to its buffer of instances stored during the warning level of DDM, which is used for training the new population in the first iteration after a drift detection. Therefore, based on these results, the DESDD-DDM version is used for comparison with baselines in the next section.

\subsection{Comparison between DESDD and Baselines}

In this last series of experiments, our objective is to verify the impact of using diverse populations of ensembles and DES to detect concept drift. DESDD is compared to DDD, DDM and Leveraging Bagging (LB). All methods were configured with Hoeffding Tree as base learning algorithm. For DESDD, 11 homogeneous ensembles were used, all with 10 Hoeffding Trees. The range for the values $\lambda$ was the same used in the previous experiment. For the baselines, they were employed using their standard versions, i.e., LB: 10 Hoeffding Trees, $\lambda=6$ and ADWIN as drift detector; DDM: 1 Hoeffding Tree; DDD: EDDM (Early Drift Detection Method) \cite{BaenaGarc2005EarlyDD} as drift detector and a population limited to 4 ensembles, all with 25 components. However, DDD was employed using ensembles with Hoeffding Trees, instead of the standard Naive Bayes, to a fair comparison. Since only one replication of the baselines was performed due to the fact that these methods have no random component, the results of only one replication of DESDD are compared in this section.

The Wilcoxon statistical test was applied by testing the equality between pairs of methods. The confidence level was 95\% ($p$-value of 0.05). Each dataset investigated was divided into batches of $30$ iterations and prequential accuracies of each batch were used to provide different values for the Wilcoxon statistical test. Table \ref{tab_metodos_comp_sin} reports classification accuracy achieved by all four methods and their execution times in seconds. 

This table shows that, except for the Agrawal Abrupt and the SEA Gradual Noise datasets, where DDM outperformed DESDD, DDD and LB in terms of classification accuracy, according to the Wilcoxon statistical test, DESDD outperformed the baselines or achieved accuracy very similar to the best rate for the remaining artificial datasets. Therefore, these results indicate that DESDD is effective for dealing with both abrupt and gradual changes, even when noise is added to the data. In terms of real datasets, the differences between accuracy rates obtained were much more significant. DESDD outperformed all the baselines in three of the four real datasets tested, and was tied at the best rate for the remaining KDDCup. These results are especially important because they indicate that DESDD can be very effective when used in real-world problems. If we compare the baselines, our results confirm the literature that DDM is better in artificial datasets, while LB achieves higher accuracy in real datasets \cite{Bifet2010a}.

% DETECÇÕES - 
\begin{table*}[htbp]
\caption{Detection test for artificial and real datasets. (D) Amount of detected drift; (FD) Amount of False Detection; (MD) Miss detection rate; (ADR) Average Delay Rate. $ADR =  \frac{\sum Delay}{NumChanges}.$ N: Noise }
\begin{center}
\begin{tabular}{|l|c|c|c|c|c|c|c|c|c|c|c|c|c|c|c|c|}
	\hline
	 & \multicolumn{4}{c|}{\textbf{DESDD}} & \multicolumn{4}{c|}{\textbf{DDD}} & \multicolumn{4}{c|}{\textbf{DDM}} & \multicolumn{4}{c|}{\textbf{LB}}  \\	
	\cline{2-17}
	\multirow{-2}{*}{\textbf{Dataset}} & \textbf{D} & \textbf{FD} & \textbf{MD} & \textbf{ADR} & \textbf{D} & \textbf{FD} & \textbf{MD} & \textbf{ADR} & \textbf{D} & \textbf{FD} & \textbf{MD} & \textbf{ADR}& \textbf{D} & \textbf{FD} & \textbf{MD} & \textbf{ADR}  \\ \hline
    Agrawal Abrupt & 9  & 2  & 2  & 93.11  & 36  & 28  & 1  & 61.0  & 6  & 1  & 4  & 41.56  & 77  & 68  & 0  & 123.56   \\ \hline 
    Agrawal Abrupt N & 5  & 0  & 4  & 94.11  & 27  & 20  & 2  & 97.11  & 12  & 8  & 5  & 50.22  & 60  & 51  & 0  & 80.89   \\ \hline 
    Agrawal Gradual & 8  & 2  & 3  & 197.11  & 22  & 16  & 3  & 55.33  & 4  & 1  & 6  & 31.78  & 70  & 61  & 0  & 73.78   \\ \hline 
    Agrawal Gradual N & 6  & 1  & 4  & 77.89  & 23  & 16  & 2  & 189.56  & 8  & 1  & 2  & 118.11  & 60  & 52  & 1  & 56.89   \\ \hline 
    Gauss & 9  & 0  & 0  & 29.0  & 30  & 22  & 1  & 22.89  & 10  & 1  & 0  & 47.56  & 44  & 35  & 0  & 34.67   \\ \hline 
    SEA Abrupt & 9  & 2  & 2  & 106.56  & 10  & 3  & 2  & 98.11  & 8  & 1  & 2  & 108.44  & 22  & 17  & 4  & 86.22   \\ \hline 
    SEA Abrupt N & 7  & 1  & 3  & 219.78  & 41  & 32  & 0  & 141.0  & 6  & 1  & 4  & 194.78  & 5  & 2  & 6  & 163.56   \\ \hline 
    SEA Gradual & 5  & 1  & 5  & 42.67  & 11  & 4  & 2  & 209.33  & 8  & 4  & 5  & 82.0  & 14  & 11  & 6  & 40.0   \\ \hline 
    SEA Gradual N & 6  & 0  & 3  & 246.44  & 29  & 23  & 3  & 87.0  & 5  & 0  & 4  & 199.67  & 0  & 0  & 9  & 0.0   \\ \hline 
    Sine1 & 9  & 0  & 0  & 15.11  & 23  & 14  & 0  & 21.0  & 16  & 7  & 0  & 19.0  & 45  & 36  & 0  & 27.56   \\ \hline \hline
	Forest Covertype & 173 & - & - & - & 2625 & - & - & - & 4301 & - & - & - & 2537 & - & - & - \\ \hline
	KDDCup & 46 & - & - & - & 10 & - & - & - & 49 & - & - & - & 45 & - & - & - \\ \hline
	Poker-Hand & 667 & - & - & - & 4798 & - & - & - & 1046 & - & - & - & 4360 & - & - & - \\\hline
	Spam & 4 & - & - & - & 18 & - & - & - & 40 & - & - & - & 10 & - & - & - \\ \hline\hline
	Total & 963 & 9 & 26 & 1121.78 & 7703 & 178 & 16 & 982.33 & 5519 & 25 & 32 & 893.12 & 7349 & 333 & 26 & 687.13 \\ \hline
	
\end{tabular}
\label{tab_metodos_deteccao}
\end{center} 
\end{table*}

Moreover, the investigated methods can be ordered with respect to computational complexity: DDM, LB, DDD and DESDD. This result was expected, since DDM employs only one classifier, LB uses an ensemble of 10 classifiers, and DDD works with a population limited to 4 ensembles (with 25 components each), while DESDD generates a population with 11 ensembles, each composed of 10 members. On the other hand, we believe that DESDD establishes a good trade-off between accuracy and complexity, since designing classifiers with a high level of generalization performance is the main objective in learning problems, especially in real-world applications. In addition, as discussed in the next paragraphs, DESDD may reduce the computational complexity involved when updating the system after a drift detection due to its higher robustness to false alarms and its precision at signaling correct drift detections.   

Besides the competitive classification accuracy rates achieved by DESDD, our method outperformed the baselines taking into account important aspects in practical tasks involving concept drift, such as: miss detection and false detection rates. In order to identify and analyze in details these aspects, Table \ref{tab_metodos_deteccao} shows values of four metrics: detected drifts; false detection; miss detection; and late detection. The last row refers to total counts of each measure. It is possible to observe in this table that DESDD significantly outperformed its competitors in terms of precisely detecting concept drifts with fewer missed or false detections, regardless of drift types. DESDD achieved the lowest false detection rate, i.e. it was the most robust method to false alarms and the most precise on signaling changes. This is an important measure in the context of practical problems involving concept drift, since false detection may lead to unnecessary system updates and higher computational cost. 

In terms of miss detection, DDD reached the lowest rate, while DESDD and LB were tied at the second best rate. However, it is very probably that the low detection rates achieved by DDD and LB are due to the fact that these two methods trigger significantly much more false alarms than DESDD (as seen in the FD columns). For instance, LB was the only method that missed all drifts in the SEA Gradual Noise dataset. This can also be the reason for the lower detection delays (ADR) attained by these two methods. These results show that DESDD can accurately detect almost all drifts, leading to faster recovery. Taking into account that DESDD is able to precisely detect drifts the earliest possible and it is robust to false alarms, it is expected to adapt faster to new concepts and provide higher accuracy. The results attained in the real datasets corroborate these observations, since DESDD, except for the KDDCup dataset, triggers significantly much less drift detection in real datasets than the other methods. As a result, DESDD achieved classification accuracy significantly superior when compared to baselines in these datasets, as seen in Table \ref{tab_metodos_comp_sin}.

\begin{figure*}
  \centering
  \subfigure[Agrawal Abrupt]{\includegraphics[width=0.32\textwidth]{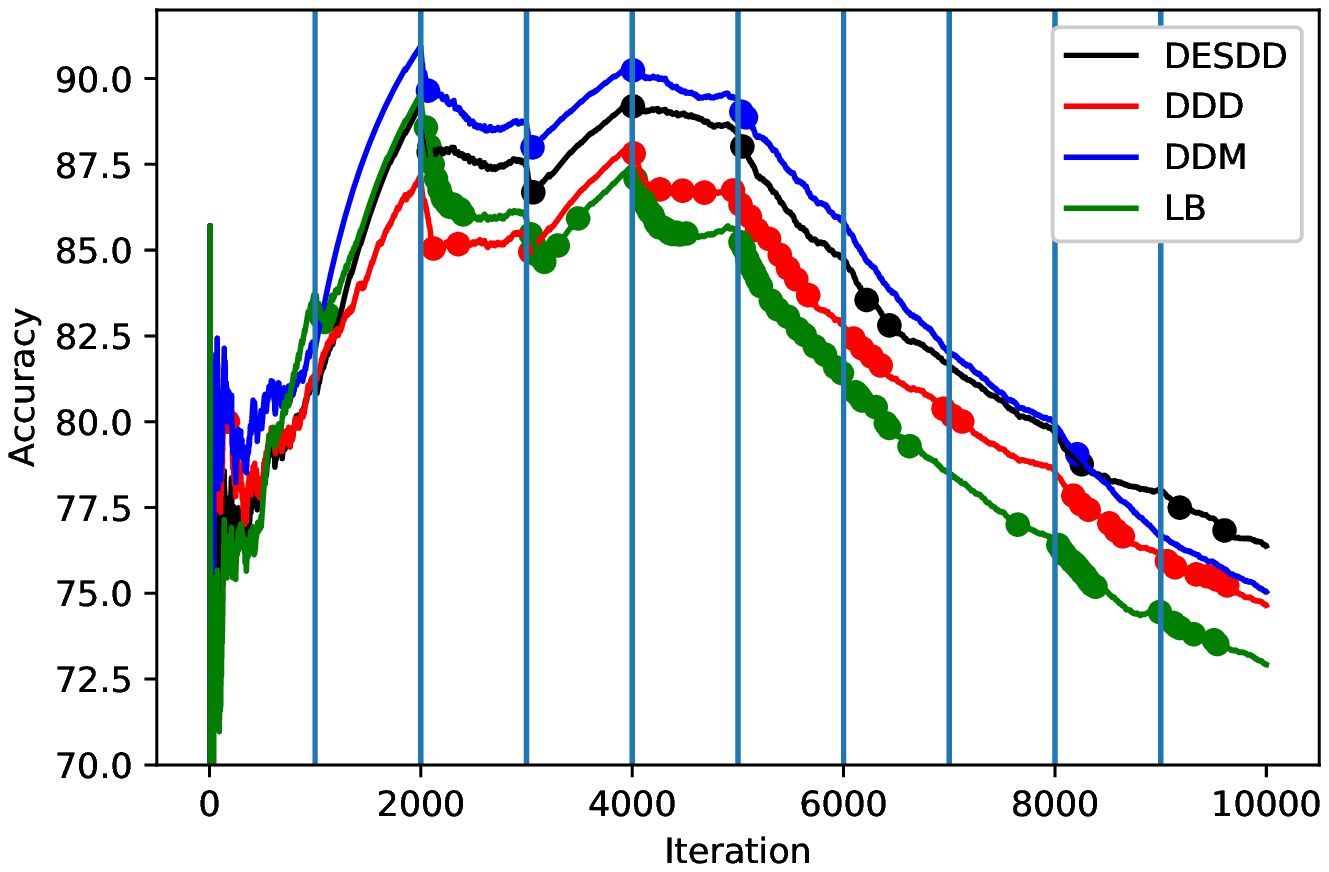}\label{AgrawalAbrupt}}
  \subfigure[SEA Abrupt]{\includegraphics[width=0.32\textwidth]{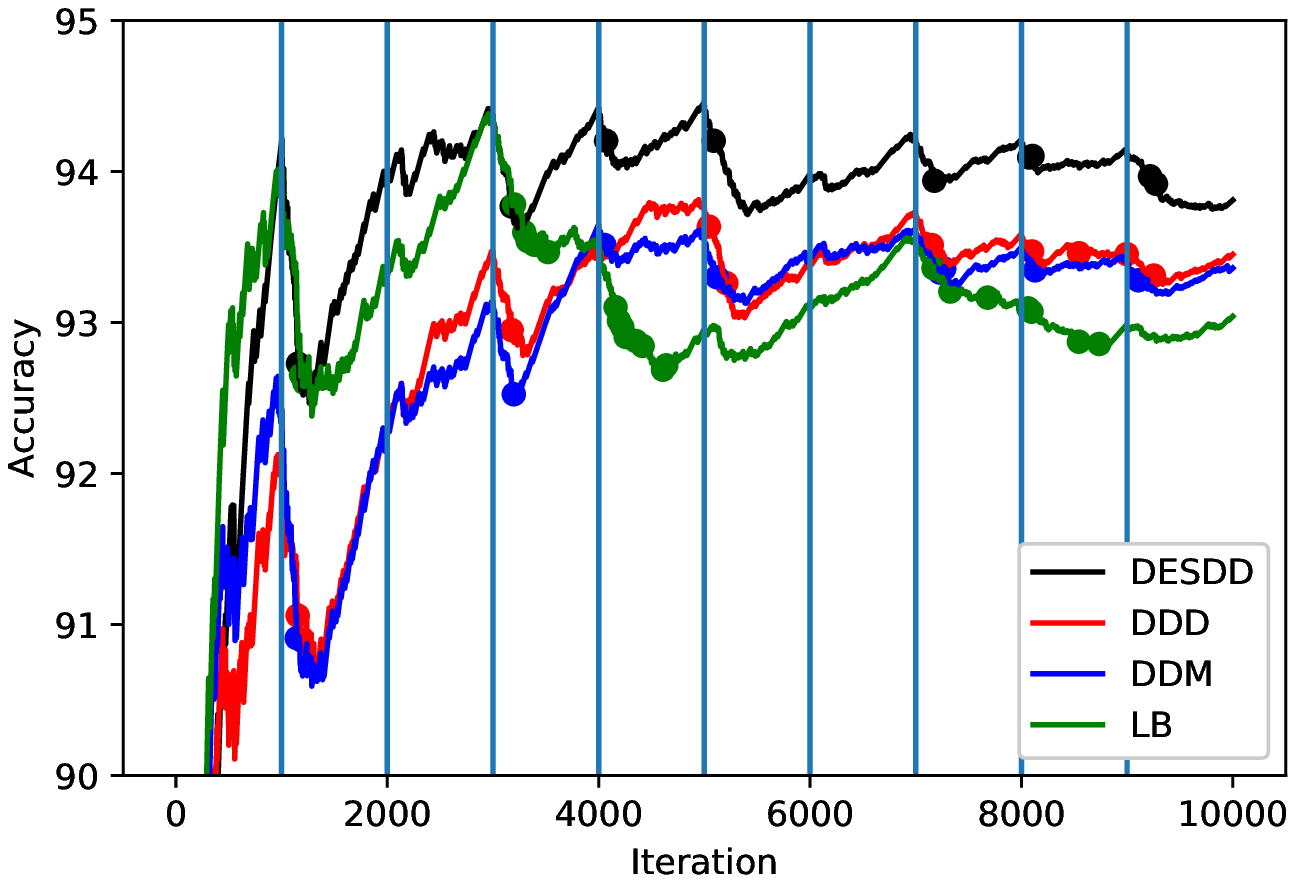}\label{SEAAbrupt}}
  \subfigure[Gauss]{\includegraphics[width=0.32\textwidth]{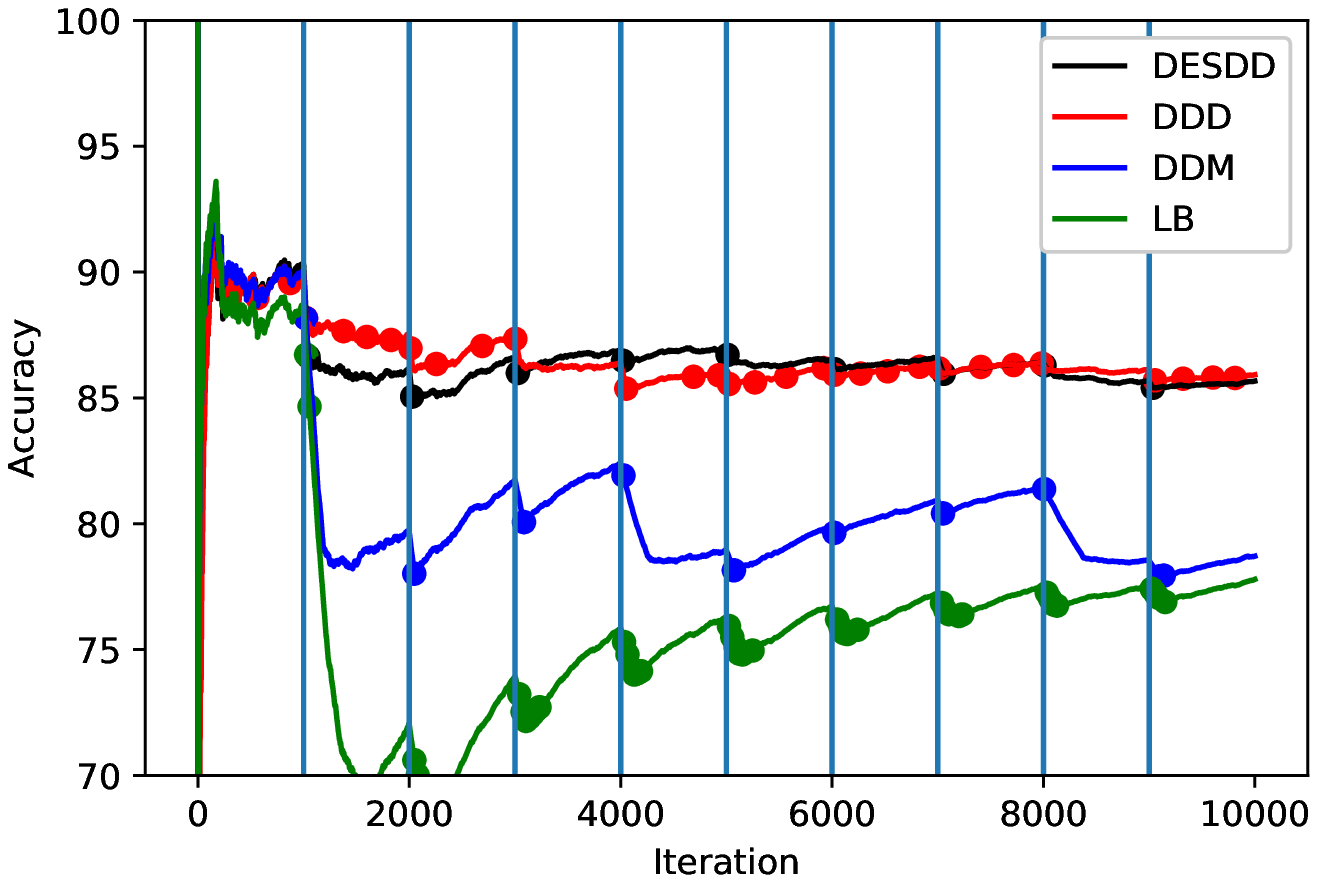}\label{Gauss}}
  \subfigure[Agrawal Gradual]{\includegraphics[width=0.32\textwidth]{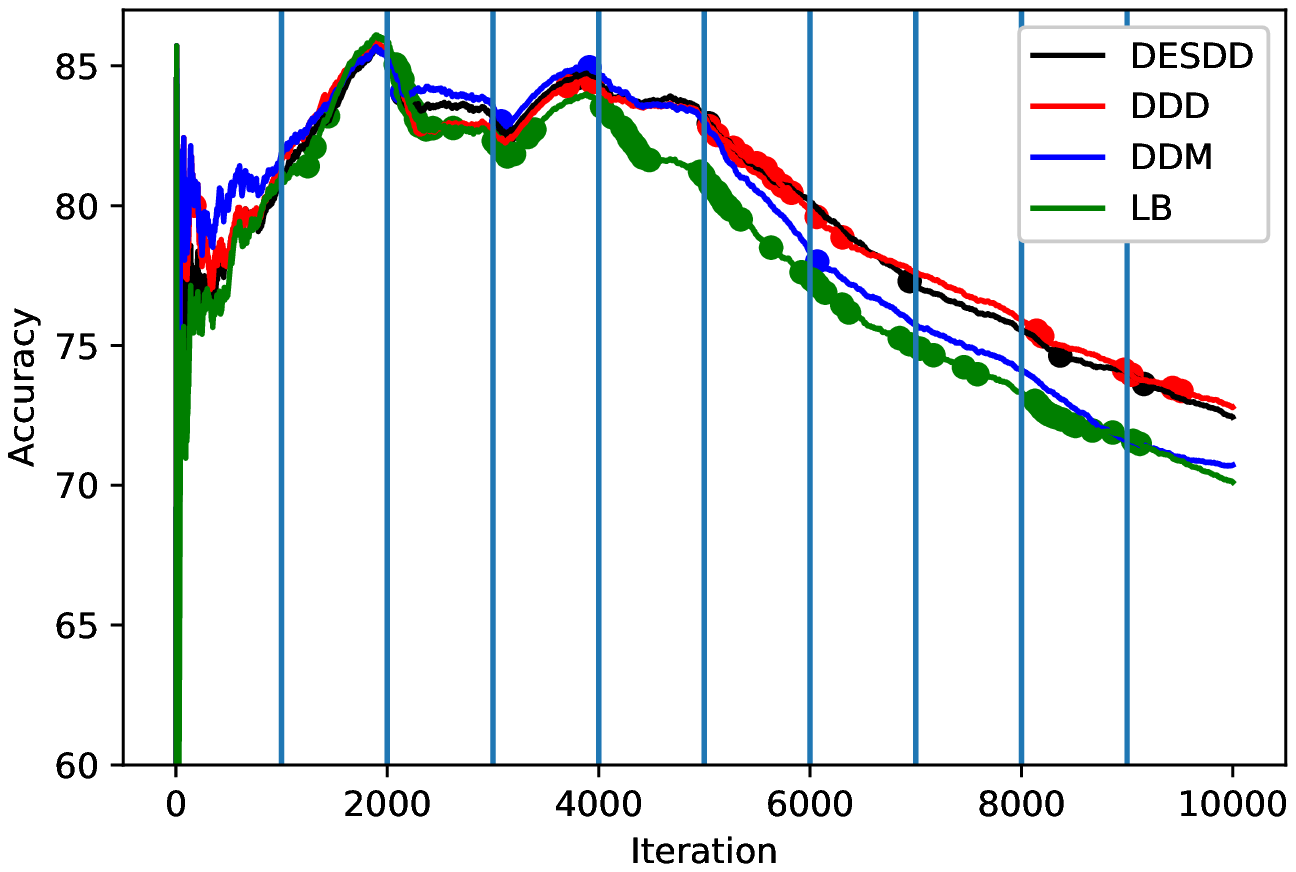}\label{AgrawalGradual}}
  \subfigure[Sine1]{\includegraphics[width=0.32\textwidth]{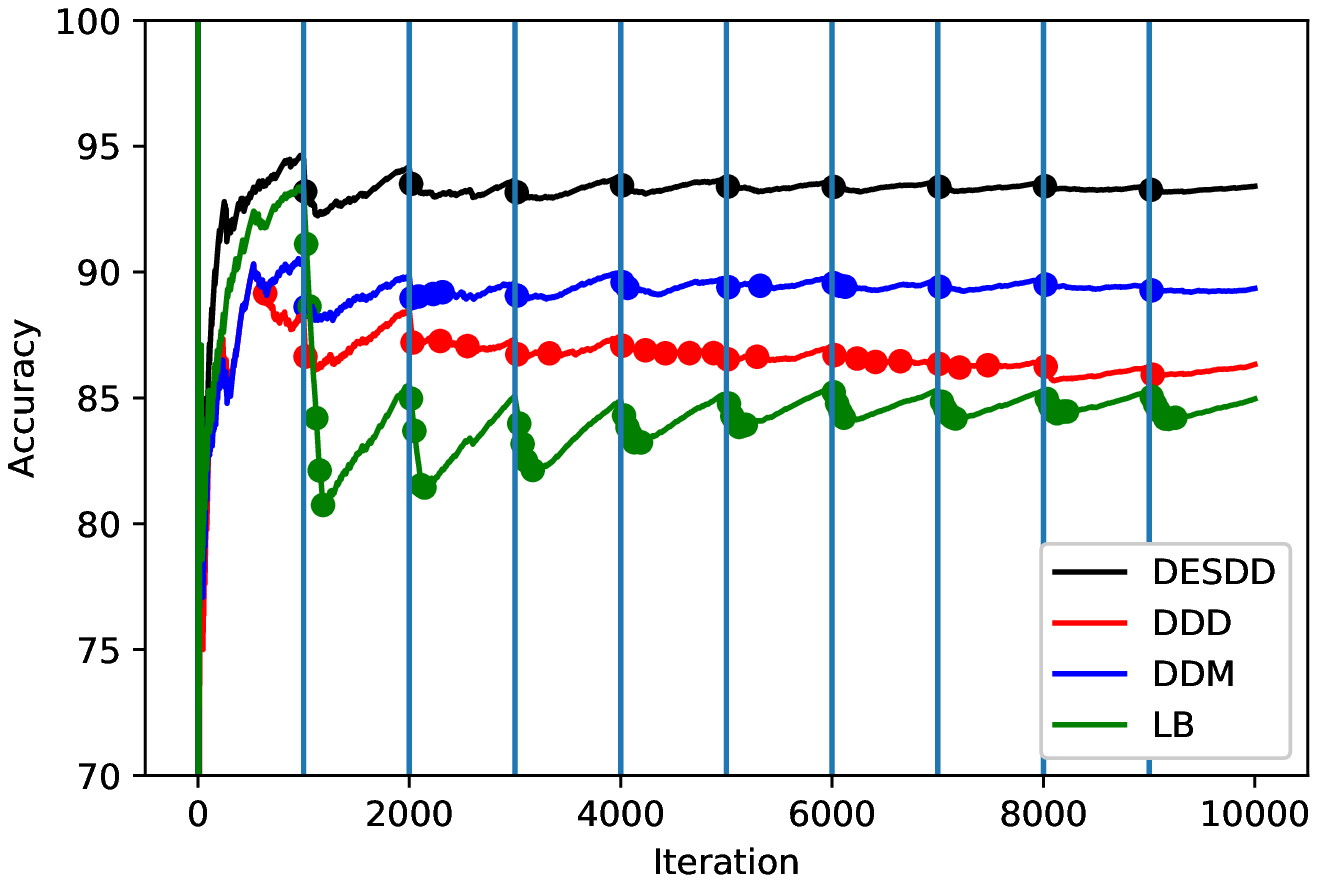}\label{Sine1}}
  \subfigure[SEA Gradual Noise]{\includegraphics[width=0.32\textwidth]{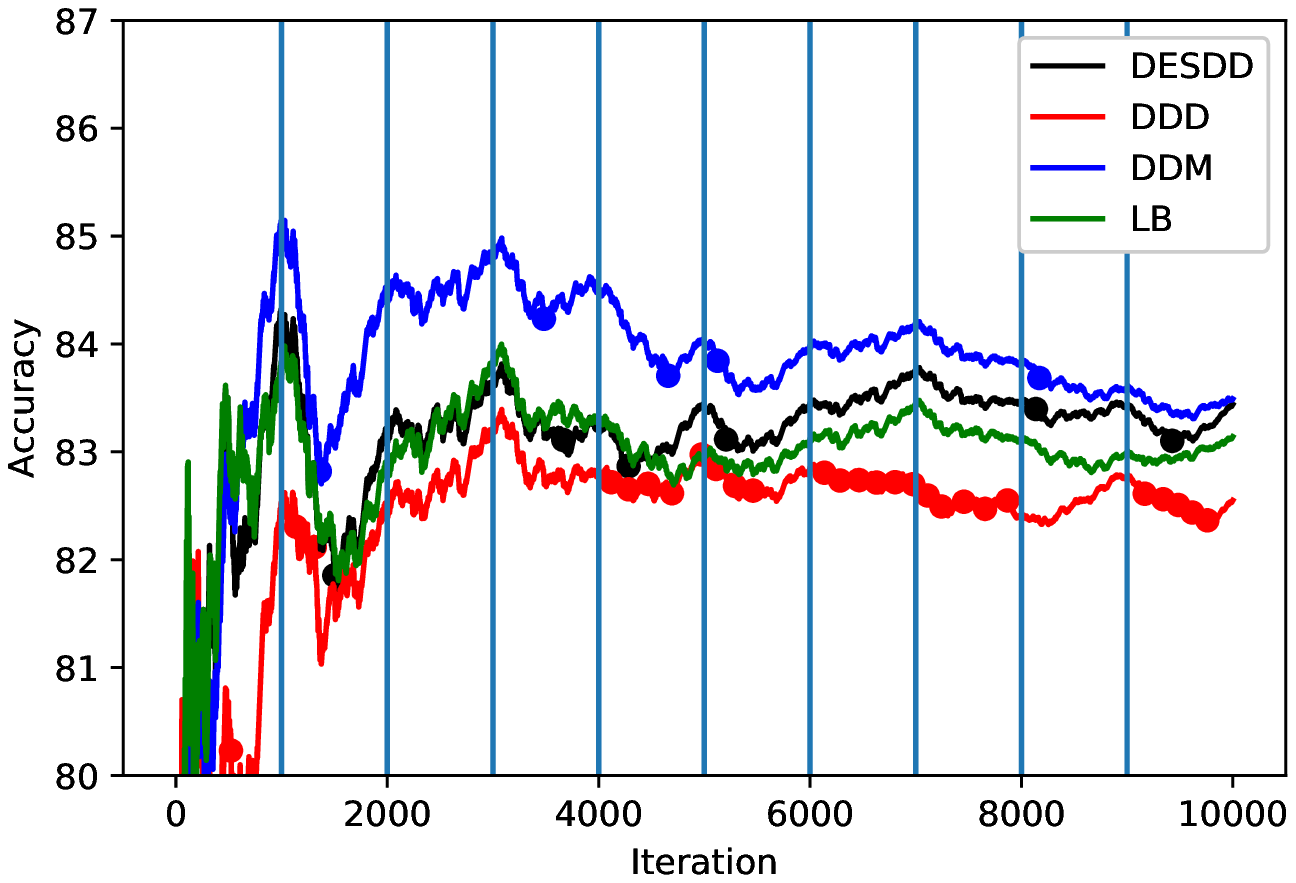}\label{SEAGradualNoise}}
  \caption{Iteration x accuracy attained by DESDD, DDD, DDM and LB on two datasets with abrupt drift (\ref{AgrawalAbrupt} and \ref{SEAAbrupt}), one dataset with abrupt drift and noise (\ref{Gauss}), two datasets with gradual drift (\ref{AgrawalGradual} and \ref{Sine1}) and one dataset with gradual drift and noise (\ref{SEAGradualNoise}). Circle indicates when a method detected a drift, while vertical line shows the real moment drift occurs.}
  \label{fig:bases_sinteticas}
\end{figure*}

These observations are reinforced in Figure \ref{fig:bases_sinteticas}, which shows the learning curves of the investigated methods on six artificial datasets: two datasets with abrupt drifts (figures \ref{AgrawalAbrupt} and \ref{SEAAbrupt}); one with abrupt drifts and noise (\ref{Gauss}); two with gradual drifts (\ref{AgrawalGradual} and \ref{Sine1}); and one with gradual drifts and noise (\ref{SEAGradualNoise}). As changes are known on these problems, each figure also shows the moment when a change occurs (vertical line) and the moment each method detects a change. It can be seen that DESDD achieves better detection performance, since it detects almost all known changes with lower delay rate and generates very few false detections on datasets presenting different types of concept drifts. It is worth noting that DDM may be deemed to be the second best strategy for detecting drift in artificial datasets. Taking into account that DDM is the drift detector employed by DESDD, we may conclude that DESDD inherits the advantage of DDM. However, the population of diverse ensembles and the dynamic selection process provided by DESDD add more stability and robustness to false alarms and allow classification accuracy improvement, especially in real datasets, where DESDD outperformed DDM in terms of classification accuracy, while generating much fewer drift detections.  

\section{Conclusion}
\label{sec:conclusion}
A novel method for drift detection is proposed in this paper. DESDD is focused on generating a diverse population of ensembles and on dynamic ensemble selection. Our experiments show that DESDD was superior to baselines in most of the tests conducted. Based on these results, it is possible to highlight that by using a population of classifier ensembles generated with different levels of diversity and by selecting classifier ensembles dynamically using a global-oriented criterion, we may improve the effectiveness of drift detectors by reducing their detection delay, false detection and miss detection rates, besides increasing their accuracy, especially in real-world application problems. Future work includes using diversity as drift detection criterion. %Finally, the general accuracy of these methods may also be increased, as well as their robustness to noise. 
%As a consequence, when these factors are reduced, drift reaction is performed only if there is changing occurrence, avoiding unnecessary system updates and reducing  computational cost. 
%\section*{References}


\begin{thebibliography}{00}
\bibitem{Brzezinski2016} Dariusz Brzezinski and Jerzy Stefanowski, ``Ensemble Diversity in Evolving Data Streams,'' pp. 229--244, 2016, Discovery Science. DS 2016. Lecture Notes in Computer Science, vol 9956.
\bibitem{BARROS2019213} Roberto S. M. de Barros and Silas G. T. C. Santos, ``An overview and comprehensive comparison of ensembles for concept drift,'' Information Fusion, vol. 52, pp. 213 - 244, 2019.
\bibitem{Zliobaite2016} I. Zliobaite, M. Pechenizkiy and J. Gama, ``An Overview of Concept Drift Applications,'' in N. Japkowicz, and J. Stefanowski (Eds.), Big Data Analysis: New Algorithms for a New Society (pp. 91-114), (Studies in Big Data; Vol. 16) Cham: CSpringer International Publishing.
\bibitem{Bifet2010a} Albert Bifet, Geoff Holmes, and Bernhard Pfahringer, ``Leveraging bagging for evolving data streams,'' Lecture Notes in Computer Science, vol. 6321 LNAI, pp. 135--150, 2010.
\bibitem{Oza2001} N.C. Oza and S. Russell, ``Experimental Comparisons of Online
and Batch Versions of Bagging and Boosting,'' pp. 359-364, 2001, ACM SIGKDD
Int. Conf. Knowledge Discovery and Data Mining.
\bibitem{Bifet2007} Albert Bifet and Ricard Gavald{\`a}, ``Learning from Time-Changing Data with Adaptive Windowing,'' pp. 443--448, 2007, 2007 SIAM International Conference on Data Mining.
\bibitem{Gama2004} J. Gama and G. Castillo, ``Learning with local drift detection,'' pp. 286–295, 2004, Advances in Artificial Intelligence. Lecture Notes in
Computer Science, vol. 3171.
\bibitem{Minku2012} L. L. Minku and X. Yao, ``DDD: A New Ensemble Approach for Dealing with Concept Drift,'' IEEE Transactions on Knowledge and Data Engineering, vol. 24, pp. 619-633, 2012.
\bibitem{ALMEIDA201867} Paulo R.L. Almeida, Luiz S. Oliveira, Alceu S. Britto and Robert Sabourin, ``Adapting dynamic classifier selection for concept drift,'' Expert Systems with Applications, vol. 104, pp. 67 - 85, 2018.
\bibitem{CRUZ2018195} Rafael M.O. Cruz, Robert Sabourin and George D.C. Cavalcanti, ``Dynamic classifier selection: Recent advances and perspectives,'' Information Fusion, vol. 41, pp. 195 - 216, 2018.
\bibitem{BaenaGarc2005EarlyDD} Manuel Baena-Garc and Jos{\'e} del Campo {\'A}vila and Albert Bifet and Ricard Gavald and Rafael Morales-Bueno, ``Early Drift Detection Method,'' pp. 77-86, 2006, Fourth ECML PKDD Int’l Workshop Knowledge Discovery from Data Streams (IWKDDS ’06).
\end{thebibliography}
\end{document}